\documentclass[11pt]{article}

\usepackage[preprint]{acl}

\usepackage{times}
\usepackage{latexsym}
\usepackage{tablefootnote}
\usepackage[T1]{fontenc}

\usepackage[utf8]{inputenc}

\usepackage{microtype}

\usepackage{inconsolata}

\usepackage{graphicx}
\usepackage{booktabs}
\usepackage{url}
\usepackage{comment}
\usepackage{multirow}
\usepackage{adjustbox}

\def\langlabel#1{\texttt{#1}}

\let\origfootnote\footnote
\renewcommand{\footnote}[1]{\kern.1em\origfootnote{#1}}
\newcommand{\punctfootnote}[1]{\kern-.1em\origfootnote{#1}}

\title{OpenLID-v3: Improving the Precision of Closely Related Language Identification -- An Experience Report}

\author{Mariia Fedorova \qquad Nikolay Arefyev \qquad Maja Buljan \qquad Jindřich Helcl  \\ {\bf Stephan Oepen} \qquad {\bf Egil Rønningstad} \qquad {\bf Yves Scherrer} \\
         Language Technology Group \\Department of Informatics \\ University of Oslo \\ \texttt{mariiaf@ifi.uio.no}}

\begin{document}
\maketitle
\begin{abstract}
    Language identification (LID) is an essential step in building high-quality multilingual datasets from web data. 
    Existing LID tools (such as OpenLID or GlotLID) often struggle to identify closely related languages and to distinguish valid natural language from noise, which contaminates language-specific subsets, especially for low-resource languages.
    In this work we extend the OpenLID classifier by adding more training data, merging problematic language variant clusters, and introducing a special label for marking noise.
    We call this extended system OpenLID-v3 and evaluate it against GlotLID on multiple benchmarks.
    During development, we focus on three groups of closely related languages (Bosnian, Croatian, and Serbian; Romance varieties of Northern Italy and Southern France; and Scandinavian languages)
    and contribute new evaluation datasets where existing ones are inadequate.
    We find that ensemble approaches improve precision but also substantially reduce coverage for low-resource languages.
\end{abstract}

\section{Introduction}

Growing interest in large-scale LLM pre-training data for languages other than English puts a spotlight on robust and broad-coverage language identification (LID).
Common pre-training datasets are typically distilled from massive collections of web documents, which are characterized by immense diversity in, for example, genres and domains, degrees of (in)formality, juxtaposition of language and non-language content, presence of machine-generated content, code switching, and other sources of variation.
For instance, the two largest and linguistically broadest pre-training datasets, FineWeb~2 \citep{penedo2025fineweb2pipelinescale} and HPLT~3.0 \citep{oepen2025hplt30largescalemultilingual}, apply LID as a document-level classification task. 
Specifically, FineWeb builds on the third-party GlotLID classifier \citep{kargaran-etal-2023-glotlid}, which supports some 2,000 distinct languages, whereas HPLT has developed a custom classifier dubbed OpenLID \citep{burchell-etal-2023-open} covering around 200 languages; see Section~\ref{sc:lid} for further background.
In this work, we seek to shed more light on LID performance and challenges in the realm of noisy web documents, with a particular emphasis on selected groups of closely related languages.

The contributions of this paper are as follows:

\begin{itemize}
    \item We train a new version of the fully open-source OpenLID system 
    for our experiments and publicly release it as OpenLID-v3.\punctfootnote{\url{https://github.com/hplt-project/openlid}} This version covers 194  languages plus `not-a-language' class.
    \item We evaluate OpenLID-v3 on mainstream LID benchmarks such as FLORES+ and UDHR and show that they are not sufficient for evaluating the quality of similar LID.\punctfootnote{\url{https://github.com/hplt-project/openlid-v3-evaluation}} In addition, we employ several existing benchmarks for similar languages and create new ones for the BCMS macrolanguage, and for Norwegian Bokmål and Nynorsk.

    \item We also report negative results on our efforts with two-step coarse-to-fine classification approach, to be found in Appendix~\ref{app:cascade}.
\end{itemize}

\section{Related Work}
\label{sc:lid}

Accurate language identification (LID) is essential for building high-quality multilingual datasets,
since documents or segments assigned to an incorrect language can severely contaminate the
language-specific subsets, especially for low-resource languages. 

\subsection{LID Tools and Methods}

Numerous LID systems have been developed over the years, including langid.py \citep{lui-baldwin-2012-langid}, Google's Compact Language Detector 2 (CLD2) and its neural network successor CLD3, the HeLI method \citep{jauhiainen-etal-2016-heli}, LanideNN \citep{kocmi-bojar-2017-lanidenn}, or AfroLID \citep{adebara-etal-2022-afrolid}.

In recent years, classifiers based on the fastText model \citep{joulin-etal-2017-bag}
have become the de facto standard in large-scale corpus processing due to their efficiency
and strong performance across many languages. FastText calculates a hidden representation of a text by summing the embeddings of words and character n-grams from this text. This representation is passed through a linear layer for classification. This approach takes both word- and character-level features into account, making it especially efficient for languages with rich morphology.

In this work, we focus on two fastText-based systems: \textbf{GlotLID} \citep{kargaran-etal-2023-glotlid} and \textbf{OpenLID} \citep{burchell-etal-2023-open}, both designed for multilingual scenarios and widely adopted in massive-scale data curation pipelines such as FineWeb \citep{penedo2024the,penedo2025fineweb2pipelinescale} or HPLT \citep{de-gibert-etal-2024-new,burchell-etal-2025-expanded,oepen2025hplt30largescalemultilingual}. OpenLID and GlotLID differ primarily in their language coverage and training data selection.
GlotLID focuses on maximizing the language coverage and supports over 2,000 languages, though it incorporates training data from sources with more restrictive licenses.
This extensive coverage results in a larger model size, but makes it suitable for use cases involving low-resource languages. In contrast, OpenLID prioritizes fully open-source training data with permissive licenses. The original version (v1) covering 201 languages was later updated (v2) to a reduced coverage of 189 languages. The reduction excluded three problematic languages and consolidated certain language varieties under their macrolanguage labels for compatibility with the FLORES+ benchmark.

\subsection{Broad-Coverage Evaluation Benchmarks}

The most widely adopted multilingual language identification benchmarks are FLORES+ and UDHR. \textbf{FLORES+}\footnote{\url{https://huggingface.co/datasets/openlanguagedata/flores_plus}} \citep{maillard-etal-2024-findings,dale-etal-2025-findings} is a
parallel corpus based on FLORES-200 \citep{nllb2022}, originally developed for machine translation evaluation. It features two disjoint publicly available data splits, `dev' and `devtest', and currently covers 225 language varieties. In our experiments, we use the former for validation and the latter for evaluation. The \textbf{UDHR} dataset consists of translations of the Universal Declaration of Human Rights
across 418 languages,\punctfootnote{We use the \texttt{udhr-lid} version of the dataset available at \url{https://huggingface.co/datasets/cis-lmu/udhr-lid}} which provides formal, declarative text in a single domain. The UDHR dataset features a test split only.

Additionally, \textbf{FastSpell} \citep{banon-etal-2024-fastspell} provides a benchmark specifically designed to discriminate between closely related languages in web documents, which addresses some of the limitations that stem from using relatively clean parallel corpora.  

Finally, \citet{oepen2025hplt30largescalemultilingual} present a manual inspection effort conducted in the context of HPLT 3.0 data collection.\punctfootnote{\url{https://github.com/hplt-project/release3_inspection/tree/main/annot_round1}} In that work, human annotators checked the prediction of OpenLID-v2 for web documents. They could label documents as containing unnatural language, porn, web artifacts, and incorrect LID. While the data hasn't been annotated for the correct language, 
it is possible to construct a LID evaluation dataset from the subset of correctly identified samples. Statistics of this resulting dataset, dubbed \textbf{HPLT-LID}, and evaluation results on it are to be found in the Appendix~\ref{app:hplt30}. We also performed re-annotation of incorrectly classified examples for some languages.

\subsection{Discriminating Between Similar Languages}

The VarDial workshop has a long tradition of shared tasks that focus on language identification in challenging settings, such as the discrimination of closely related languages \citep[e.g.][]{zampieri-etal-2014-report,gaman-etal-2020-report,aepli-etal-2023-findings,chifu-etal-2024-vardial}. For example, shared tasks have focused on the Bosnian-Croatian-Montenegrin-Serbian macrolanguage or on regional languages of Italy. We leverage datasets made available in this context for a more fine-grained analysis of OpenLID-v3 (see Section~\ref{sec:case-studies}).

\subsection{Evaluation Metrics}

\citet{caswell-etal-2020-language} argue that the commonly used metric precision (and its derivative, F1-score) are misleading when evaluating models on the standard LID datasets, as the models face incomparably larger language imbalance in the real web crawls compared to any reasonable labeled dataset. For example, their baseline model achieving a median F1-score of 98\% on benchmark data produced a set of monolingual corpora with a median precision of 5\% only when evaluated manually on real data. On the other hand, recall and false positive rate (FPR) are not susceptible to class imbalance, and thus are recommended for model comparison.

Since most of our benchmarks are single-label, we report FPR, precision, recall and F1-score for them. For multilabel benchmarks, which are available for some related languages, where short samples may be valid in more than one language, we also report loose (is subset) and exact match metrics \citep[cf.][]{fedorova-etal-2025-multi}.\punctfootnote{\url{https://github.com/ltgoslo/slide}} These metrics (we refer to them as multilabel classification metrics throughout the paper) are still less strict than our default ones, because they do not depend on the exact number of false positives or negatives in a set of predicted labels.

\section{OpenLID-v3}

This work is motivated by our experience with OpenLID-v2 in the context of its application to a large-scale web dataset HPLT 3.0, and the results of its manual inspection \citep{oepen2025hplt30largescalemultilingual}. We identified a number of issues with OpenLID-v2, which led us to provide an updated version, OpenLID-v3, with a slightly different language inventory. We focus on improving OpenLID-v2 (rather than GlotLID, for example) because of the permissive license of all its training data. 
The identified issues are as follows:

\begin{itemize}
    \item Support for the Latin language, which was present in OpenLID-v1, but removed in v2.

    \item OpenLID-v2 contains only Serbian in Cyrillic script, while
   Latin script is also widely used in Serbian non-governmental public and private communication.
As a result, OpenLID-v2 erroneously classifies Serbian in Latin script 
as Bosnian or Croatian; according to the manual inspection, half of HPLT 3.0 Bosnian turned out to be  Serbian written in Latin script.

    \item Since OpenLID-v2's class inventory is limited to 200 languages, any text that is not natural language (e.g., code, broken encoding -- we label these instances as not-a-language, using \langlabel{zxx\_Zxxx} throughout the paper) or a natural language outside of the 200 the model was trained for (we refer to such instances as \langlabel{other}), are still predicted to belong to one of the existing classes. In HPLT 3.0, the problem of not-a-language classes was overcome by rule-based filtering after LID, and the problem of the \langlabel{other} class was solved by thresholding the predictions by 0.5 softmax scores. However, some classes can still accumulate a large number of not-a-language and \langlabel{other} documents. We call this the \emph{trash bin phenomenon}; in HPLT 3.0, Ligurian was found to be such a `trash bin' language \citep{oepen2025hplt30largescalemultilingual}.

    \item Some highly similar languages showed high 
confusion 
when tested on the FLORES+ development split and UDHR (Arabic dialects, Persian languages, Bambara and Dyula, Dzongkha and Tibetan, and other languages listed in Table \ref{tab:appendix-data-sources}).

\end{itemize}

\noindent Finally, the amount of identified data for certain languages in the HPLT dataset is very small compared to its number of speakers, e.g. Bengali and Tamil. See Table \ref{tab:appendix-data-sources} for the full list of such languages.

When developing OpenLID-v3, we approached these issues in the following ways:

\begin{itemize}
    \item We merged 8 Arabic dialects into one Arabic macrolanguage: \langlabel{ara\_Arab}; and two Persian varieties: \langlabel{pes\_Arab} and \langlabel{prs\_Arab} into the Farsi macrolanguage \langlabel{fas\_Arab}. This follows the OpenLID-v1 approach. Likewise, we merged Bambara and Dyula, which are mutually intelligible \citep{bambara}.
    \item We introduced the not-a-language class \langlabel{zxx\_Zxxx}, using the GlotLID training data labeled as \langlabel{und\_*} and \langlabel{zxx\_*}. 
    \item We extended the training data for several languages, including Latin, and Serbian written in Latin script. We relied on the subsets of the GlotLID training data which were reported by its authors to not be noisy. We also removed one possibly noisy OpenLID-v2 training subset and added the most recent Wikipedia dumps, where adding GlotLID training data showed insufficient improvements during validation. The affected languages and data sources used are summarized in Table \ref{tab:appendix-data-sources} in the Appendix. 
    \item We also experimented with ensembling OpenLID-v3 and GlotLID by top-1 and top-3 agreement. Top-3 agreement worsened the results, which is expected for a single-label model, where only one class gets a relatively high softmax score, and all other classes are a long tail of small numbers.
\end{itemize}

The list of language labels supported by OpenLID-v3 is given in Table \ref{tab:openlidv3-langs} in Appendix \ref{app:openlidv3-languages}.

\begin{figure}[t]
    \centering
    \includegraphics[width=0.5\textwidth]{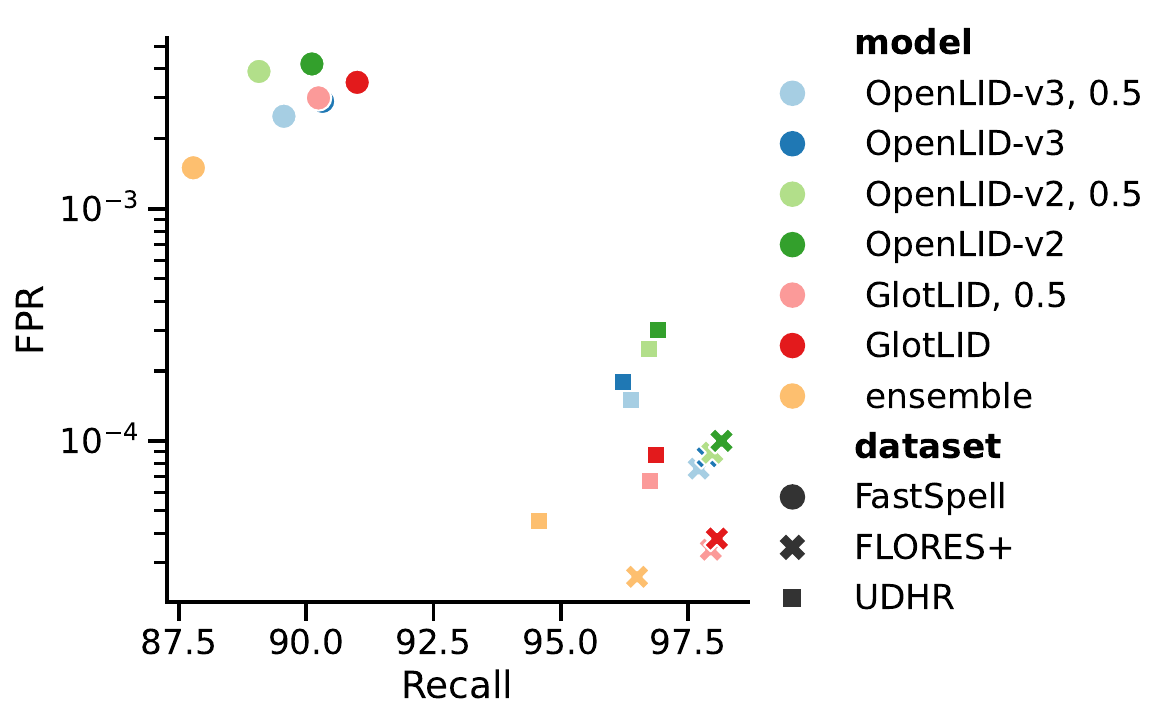} 
    \caption{Model comparison on three LID benchmarks, FPR (lower is better) vs. recall (higher is better). For the full set of metrics see Table~\ref{tab:many}. FLORES+ refers to the devtest split, and FastSpell refers to our version, which excludes Nynorsk.}
    \label{fig:multiling-banchmarks}
\end{figure}

\subsection{Results}

Figure~\ref{fig:multiling-banchmarks} compares the results of our OpenLID-v3 model with the previous version (OpenLID-v2) as well as with GlotLID, focusing on the popular LID benchmarks: FLORES+, UDHR, and FastSpell. Following HPLT 3.0, we  experimented wth applying the same softmax thresholding of 0.5 to all three models. 

OpenLID-v3 is on par with both OpenLID-v2 and GlotLID; the difference with and without thresholding softmax scores is not large among models, but the results are slightly better for OpenLID-v3 with thresholding, and for GlotLID without. This different behavior may be explained by the fact that GlotLID's `trash bins' fall outside of the classes we evaluate on. In our further experiments, we use OpenLID-v3 with thresholding, and GlotLID without, unless otherwise specified, including the ensembling approach.

Top-1 ensembling resulted in the lowest FPR across all datasets. Based on our experiments, this approach was used to produce HPLT 4.0 datasets.\punctfootnote{\url{https://hplt-project.org/datasets/v4.0}}

\section{Case Studies on Related languages} \label{sec:case-studies}

Results of both OpenLID and GlotLID are high on multilingual benchmarks. However, lower precision for similar languages may be hidden when averaging their scores with those of languages that are easy to distinguish. For this reason, we perform additional evaluation for groups of closely related languages. We choose three groups of such languages, within those with updated training data, based on benchmark availability and our language expertise.

\begin{table*}
\centering
\resizebox{\linewidth}{!}{%
\scriptsize
\begin{tabular}{lccc|ccc|ccc}
    \toprule
    \textbf{Model} & \multicolumn{3}{c}{\textbf{Bosnian}} & \multicolumn{3}{c}{\textbf{Croatian}} & \multicolumn{3}{c}{\textbf{Serbian Latin}} \\
    & FPR & Prec. & Rec. & FPR & Prec. & Rec. & FPR & Prec. & Rec. \\
    \midrule
    \textit{Twitter users} & \multicolumn{3}{c}{n=21} & \multicolumn{3}{c}{n=22} & \multicolumn{3}{c}{n=94} \\
    \midrule
    OpenLID-v3 & \textbf{0.23} & \textbf{37.21} & 76.19 & 0.026 & 80.00 & 54.54 & \textbf{0} & \textbf{100.00} & \textbf{30.85} \\ 
    OpenLID-v2 & 0.8 & 16.96 & \textbf{90.48} & \textbf{0.017} & \textbf{84.62} & 50.00 & 0 & 0 & 0 \\
    GlotLID & 0.42 & 16.95 & 47.62 & 0.26 & 36.17 & \textbf{77.27} & 0.12 & 82.14 & 24.49 \\
    Ensemble & 0 & 0 & 0 & 0 & 0 & 0 & 0 & 0 & 0 \\
    \midrule
    \textit{ParlaSent} & \multicolumn{3}{c}{n=153} & \multicolumn{3}{c}{n=1,387} & \multicolumn{3}{c}{n=1,060} \\
    \midrule
    OpenLID-v3 & 0.22 & 11.59 & 47.71 & 0.13 & 87.49 & 79.16 & 0.019 & 95.00 & \textbf{53.77} \\ 
    OpenLID-v2 & 0.44 & 0.08 & \textbf{57.51} & 0.26 & 78.05 & 80.25 & 0 & 0 & 0 \\
    GlotLID & 0.26 & 9.59 & 43.79 & 0.242 & 81.02 & \textbf{90.48} & 0.006 & 97.10 & 31.60 \\
    Ensemble & \textbf{0.14} & \textbf{13.15} & 34.64 & \textbf{0.11} & \textbf{89.25} & 77.21 & \textbf{0.003} & \textbf{98.38} & 28.77 \\
    \midrule
    \textit{HPLT-LID (reannotated)} & \multicolumn{3}{c}{n=5} & \multicolumn{3}{c}{n=7} & \multicolumn{3}{c}{n=110} \\
    \midrule
    OpenLID-v3 & 0.15 & \textbf{5.26} & \textbf{40.00} & 0.03 & 42.86 & \textbf{85.71} & 0 & \textbf{100.00} & \textbf{50.00} \\ 
    OpenLID-v2 & 0.9 & 1.85 & \textbf{40.00} & 0.09 & 21.43 & 42.86 & 0 & 0 & 0 \\
    GlotLID & 0.4 & 0.01 & 0.2 & 0.05 & 31.58 & \textbf{85.71} & 0 & \textbf{100.00} & 3.64 \\
    Ensemble & \textbf{0.146} & 2.70 & 20.00 & \textbf{0.03} & \textbf{46.15} & \textbf{85.71} & 0 & \textbf{100.00} & 3.64 \\
    \bottomrule
\end{tabular}
}
\caption{Performance comparison across three benchmark datasets showing false positive rate (FPR), precision (Prec.), and recall (Rec.) for Bosnian, Croatian, and Serbian in Latin script. OpenLID-v3 uses softmax threshold 0.5. Ensemble represents top-1 agreement between OpenLID-v3 and GlotLID. Best values per metric and language are in \textbf{bold}.}
\label{tab:bcs-metrics}
\end{table*}

\subsection{Bosnian, Croatian, Montenegrin and Serbian}
\label{sec:bcs}

Bosnian, Croatian and Serbian are, along with Montenegrin, South Slavic languages spoken in the West Balkans; part of a group of historically and geographically close languages that are, for the most part, mutually intelligible to speakers of any one of them. They differ by variations in vocabulary and grammatical features -- most notably, the orthography of the Slavic \textit{jat} (\v e) \citep{mevsanovic2011kontrastivna,karavdic2017kako}. Montenegrin is only present in GlotLID class inventory; for this reason, we do not focus on it. The analysis of its GlotLID predictions is to be found in Appendix \ref{app:hbs}.

\paragraph{Test data.} We use three datasets for the BCMS language group.\punctfootnote{
Exact scores on these benchmarks are to be found in the Appendix~\ref{app:hbs}.} First, BCMS \textbf{Twitter user} dataset \citep{clarin_twitter}, also featured in the BENCHić BCMS benchmark \citep{rupnik-etal-2023-benchic}. We use the version with multi-label annotations \citep{miletic2024gold}, comprising roughly 123 instances (in the test set), where each instance corresponds to one user and contains many texts written by that user. Second, we use the BCS portion of \textbf{ParlaSent} \citep{Mochtak_Rupnik_Ljubešić_2023}, comprising roughly 18,000 sentences from transcriptions of parliamentary debates, annotated with the speaker's country of origin.
Third, we make use of the \textbf{HPLT-LID} data. 
Since Serbian in Latin script was absent in OpenLID-v2, pre-annotation was only performed for Bosnian and Croatian. In total, 804 sentences were pre-annotated (402 predicted to be Bosnian and 402 predicted to be Croatian). Out of them, 114 Bosnian samples and 13 Croatian were pre-annotated to be false positives. We performed re-annotation of these false positive samples to their correct labels; the majority of them was found to be Serbian. The annotation was done by a native speaker of Croatian with linguistic background. One sample was detected to be valid both in Croatian and Serbian and annotated as multilabel; we excluded it from metric calculation. 

\subsubsection{Quantitative Evaluation}

Table \ref{tab:bcs-metrics} presents FPR, precision and recall per language on Twitter user, ParlaSent and reannotated BCS part of HPLT-LID.

None of the models scored high on all three datasets. As HPLT 4.0 developers, we were mostly interested in results on Twitter, since its texts were the closest to the noisy web data, containing hyperlinks, emodjis etc. OpenLID-v3 turned out to be the best for Bosnian and Serbian on it; slightly higher precision of Croatian obtained from OpenLID-v2 comes at cost of lack of Serbian class label, which is not what we aim at. Importantly, GlotLID and OpenLID-v3 always disagree on Twitter data, which emphasizes that ensembling should be done cautiously for particular language groups. For the future work on improving predictions for the Twitter dataset, there is strong need for multilabel training data, silver approach from \citep{fedorova-etal-2025-multi} might be helpful for future work.

Scores on ParlaSent were the highest, especially in precision of Serbian, which proves it to be the `easiest' dataset.

Results of evaluation on reannotated HPLT-LID were similar to those on Twitter: OpenLID-v3 performed the best, and there were a high disagreement between it and GlotLID. However, the confusion of Bosnian and Serbian was still high.

Table 2 shows multilabel classification metrics on Twitter data. By these metrics, OpenLID-v3 is a clear winner over GlotLID.

\begin{table}[]
\centering
\resizebox{\linewidth}{!}{%
\begin{tabular}{l|ccccc}
    \toprule
\textbf{Model} & \textbf{Loose} & \textbf{Exact} & \textbf{bos} & \textbf{hrv} & \textbf{srp} \\
 & \textbf{Acc.} & \textbf{Acc.} & \textbf{F1} & \textbf{F1} & \textbf{F1} \\
    \midrule
    OpenLID-v3 & \textbf{46.34} & \textbf{40.65} & \textbf{57.14} & \textbf{72.73} & \textbf{49.15} \\ 
    GlotLID & 40.65 & 33.33 & 28.17 & 55.74 & 40.00 \\
    \bottomrule
\end{tabular}
}
\caption{Loose accuracy, exact match accuracy and loose F1s per languages on multilabel BCS test data (Twitter). Ensemble@1 never agrees (always \langlabel{other}, all metrics zero).}
\label{tab:bcs-twi}
\end{table}

\subsubsection{Common Errors}

\begin{table*}
    \small
    \centering
    \begin{tabular}{lrrrrrrrrrr}
    \toprule
\textit{(gold/predicted label)}    & \texttt{b/h} & \texttt{b/s} & \texttt{b/x} & \texttt{h/b} & \texttt{h/s} & \texttt{h/x} & \texttt{s/b} & \texttt{s/h} & \texttt{s/x} & \% of total \\
    \midrule 
NE confusion          & 12.0  &  -    & 2.0   & 4.0   & -     & 2.0   & 10.0  & 6.0   & 8.0   &  6.1 \\
lexical overlap        & \textbf{34.0}  &  -    & 20.0  & 26.0  & -     & 2.0   & \textbf{44.0}  & 12.0  & 6.0   & 20.5 \\
historic forms         & -     &  -    & -     & \textbf{36.0}  & -     & 2.0   & -     & -     & -     &  5.8 \\
\textit{da} confusion  & -     & \textbf{66.6}  & 4.0   & -     & \textbf{100.0} & -     & -     & -     & -     &  1.7 \\
ungrammatical syntax   & 10.0  &   -   & \textbf{32.0}  & 10.0  & -     & \textbf{64.0}  & 6.0   & 14.0  & \textbf{28.0}  & 22.5 \\
total ambiguity        & 26.0  & -     & 12.0  & 6.0  & -     & 20.0  & -     & \textbf{34.0}  & \textbf{26.0}  & 19.9 \\
mislabeled minority rep & -    & 33.3  & 4.0   & 2.0   & -     & -     & -     & -     & -     &  1.2 \\
(other/unknown)        & 18.0  &   -   & 4.0   & 16.0  & -     & 10.0  & \textit{40.0}  & \textit{34.0}  & \textit{32.0}  & 22.2 \\
    \bottomrule
    \end{tabular}
    \caption{V3-ensemble evaluation on ParlaSent; percentage of error occurrences, per sample of 50 mislabeled documents for each mismatched pair of labels. Exceptions are \texttt{b/s}, \texttt{b/x}, and \texttt{h/s}, which have a total of 2, 40, and 3 datapoints each, respectively. The bottom row, \textit{other}, is a catch-all for unambiguous documents with clear language markers, which were nevertheless mislabeled.}
    \label{tab:errors_bcs}
\end{table*}

We performed manual analysis of BCS predictions from all models and observed the following seven most frequent error patterns:

\paragraph{NE confusion} Named entities are a common source of confusion
. The presence of a different country name (e.g.\ Serbian news article about Croatia), an individual or institution that adheres to the other language's naming convention, or even word type confusion (\textit{``\textbf{Tome} se niko ne raduje''} (``Nobody's looking forward to that'', compared with the common Slovenian name \textit{Tome}) identified as Slovenian), leads to mislabelling, due to lack of other language markers.

\paragraph{Lexical overlap} For a human reader, the current ortography of \textit{jat} is a strong discriminator between Serbian and Bosnian or Croatian, the former prefering the \textit{e} form in the standard language (e.g.\ \textit{videti}, \textit{mleko}), and the other two using \textit{je/ije} (\textit{vidjeti}, \textit{mlijeko}). However, documents are frequently mislabeled between Bosnian and Serbian, despite the presence of clear \textit{e/ije} indicators, if the document also contains roots and noun/adjective inflections that are common to both Bosnian and Serbian, whereas Croatian uses a diferent surface form of the word (e.g.\ \textit{obavezno} (Bosnian, Serbian) vs.\ \textit{obvezatno} (Croatian)). The frequency of these error occurrences (\textit{lexical overlap} in \ref{tab:errors_bcs}) seems to indicate that common roots and shared lexemes are a stronger signal than \textit{jat} ortography.

\paragraph{Historic forms} Similarly, all models frequently mislabel ParlaSent documents in Croatian as either Bosnian or Serbian, when the speakers slip into historic lexeme forms that were part of the shared Serbo-Croatian language, but are not present in current standard Croatian. Older speakers and colloquial language still frequently use \textit{historic forms}, which again leads to mislabeling despite the unambiguous grammatical markers.

\paragraph{\textit{da} confusion} Another theoretically strong indicator is the difference between the future tense in Croatian (modal verb + infinitive, e.g.\ \textit{(ho)ću glasati}, \textit{glasat ću}) and Serbian (modal verb + \textit{da} + present simple, e.g. \textit{(ho)ću da glasam}); both forms are accepted in Bosnian. Frequently, Croatian documents are mislabelled as Serbian when there are multiple occurences of the conjuction \textit{da} (meaning \textit{that}; not to be confused with the particle \textit{da} (\textit{yes})). E.g.,\ \textit{``Ne sumnjam \textbf{da} je ovaj zakon nekima donio dobro i ne sumnjam \textbf{da} oni hvale ovaj zakon.''}\ (``I do not doubt \textbf{that} this legislation brought good to some, and I do not doubt \textbf{that} they praise it.'') -- Croatian mislabelled as Serbian -- vs.\ \textit{``\textbf{Neću da glasam} za taj zakon.''} (``I \textbf{will not / don't want to vote} for this legislation.'') -- true Serbian.

\paragraph{Ungrammatical syntax} Ungrammatical syntax comprises documents in non-standard language; mainly long run-on sentences, with a lack of punctuation, incorrect syntax, frequent interjections, and generally language that is highly colloquial in structure and word choice. While this mislabeling occurs across most pairings, these documents are most frequently labeled as \textit{not-a-language} (\langlabel{zxx\_Zxxx}), particularly when there is a lack of any distinguishing grammatical or lexical language markers.\\

Finally, some datapoints are mislabeled due to extralinguistic factors. 

\paragraph{Total ambiguity} Some documents have no clear language markers between BCS. Most of these documents are mislabeled across models, and would arguably be hard even for a human annotator and native speaker to tease apart, due to a lack of unambiguous linguistic features.

\paragraph{Mislabeled minority representative} Mislabeled minority representative is specific to the ParlaSent dataset, where statements are labeled according to the national parliament in which the discussion occurred. In a handful of cases, it is clear that the speaker is speaking a minority language, and the model has identified it correctly, though it doesn't match the parliament country of origin.

We performed a quantitative study of these error types on the ParlaSent predictions obtained from the best model (OpenLID-v3 top-1 ensemble with GlotLID). The results are presented in the Table \ref{tab:errors_bcs}. The last row (\textit{other/unknown}) quantifies unambiguous documents with clear language markers, which were nevertheless mislabeled by the model. This is a particularly frequent occurrence for Latin-script Serbian. 

\subsection{Romance Languages of Italy and France}

The HPLT 3.0 compilation efforts reported language identification issues connected to Ligurian. 31\% of the samples annotated as Occitan in the UDHR dataset were predicted as Ligurian by OpenLID-v2. Upon closer inspection, it turned out that the Occitan part of UDHR contained seven translations -- three truly Occitan ones and four Francoprovençal ones (see Table~\ref{tab:oci_frp} in the Appendix). Although Ligurian, Occitan and Francoprovençal are all Romance languages and are spoken in adjacent areas (Southern France,  Northwestern Italy and Western Switzerland), they are generally considered to belong to different genealogical sub-groups \cite{oxford-guide-romance,ramponi-tacl-italy}. Since Francoprovençal was missing from OpenLID-v2's class inventory, its instances were erroneously predicted as Ligurian.

This suggests that (a) multilingual benchmarks cannot be fully relied on when it comes to closely related, unstandardized languages, and (b) LID tools struggle to identify low-resource languages especially if higher-resourced closely related varieties are present in their training data.

We further investigate the performance of OpenLID-v3 on these Romance languages. As we were unable to find a test set that covered Francoprovençal, Ligurian and Occitan,\punctfootnote{While there exists an effort on Occitan LID \citep{miletic-scherrer-2022-ocwikidisc}, we could not use it in our experiments, as it contains silver labels only.} we use the \textbf{ITDI} dataset instead, which covers several languages and dialects of Italy \citep{2022-findings-vardial}. The metrics were only calculated for Ligurian, Venetian and Friulian, as only these three languages are predicted by both OpenLID-v3 and GlotLID.

\begin{table}
\adjustbox{max width=\linewidth}{%
    \begin{tabular}{l|rrrr}
    \toprule
       Model  & FPR & F1 & Precision & Recall \\
    \midrule
    OpenLID-v3 & 0.01 & 80.88 & 96.51 & 74.22 \\
    GlotLID & 0.01 & \textbf{82.37} & 97.60 & \textbf{75.15} \\
    ensemble & \textbf{0.004} & 76.68 & \textbf{98.47} & 68.95 \\
    \bottomrule
    \end{tabular}
}
    \caption{Comparison of OpenLID-v3 and GlotLID on the ITDI test set restricted to Ligurian, Venetian and Friulian (4,744 samples).}
    \label{tab:lij}
\end{table}

The results are presented in Table~\ref{tab:lij}. While GlotLID is better than OpenLID-v3 both in precision and recall, ensembling them achieves the best precision and the lowest FPR. For both models, Venetian confusion is the highest with Italian, but OpenLID-v3 predicts fewer Italian false positives and more not-a-language ones (which is never the case for GlotLID). Friulian is mostly predicted correctly, more often with OpenLID-v3 (1,284 true positives) than with GlotLID (1,275). Ligurian is also more often predicted correctly by both models, with Sicilian and Friulian as the top confusion instances. For OpenLID-v3, the third most frequently predicted class is not-a-language. Ensembling the two models removes 45\% of Venetian samples, while the number of Ligurian and Friulian ones remains roughly the same. 

In sum, this case study emphasizes the importance of fine-grained benchmarks for related Romance languages. GlotLID might be a good choice as a standalone model; however, the most precise results are obtained when it is ensembled with OpenLID-v3, although at the price of removing samples from other easily confused languages such as Venetian.

\begin{table*}
\centering
\resizebox{\linewidth}{!}{%
\begin{tabular}{lccc|ccc|ccc|ccc|ccc}
    \toprule
    \textbf{Model} & \multicolumn{3}{c}{\textbf{Norwegian Bokmål}} & \multicolumn{3}{c}{\textbf{Danish}} & \multicolumn{3}{c}{\textbf{Norwegian Nynorsk}} & \multicolumn{3}{c}{\textbf{Swedish}} & \multicolumn{3}{c}{\textbf{Other}} \\
    & FPR & Prec. & Rec. & FPR & Prec. & Rec. & FPR & Prec. & Rec. & FPR & Prec. & Rec. & FPR & Prec. & Rec. \\
    \midrule
    \textit{SLIDE} & \multicolumn{3}{c}{n=2,098} & \multicolumn{3}{c}{n=677} & \multicolumn{3}{c}{n=1,628} & \multicolumn{3}{c}{n=1,250} & \multicolumn{3}{c}{n=1,745} \\
    \midrule
    OpenLID-v3 & 0.04 & 88.73 & 87.04 & 0.02 & 81.81 & 83.75 & 0.03 & 89.66 & \textbf{84.64} & 0.008 & 96.11 & 96.80 & \textbf{0.03} & 89.68 & 95.13 \\ 
    OpenLID-v2 & 0.05 & 87.33 & 88.04 & \textbf{0.01} & 85.27 & 80.35 & 0.03 & 90.47 & 84.58 & 0.008 & 95.88 & 96.8 & \textbf{0.03} & 89.63 & 95.58 \\
    GlotLID & 0.047 & 88.32 & \textbf{90.47} & 0.02 & 80.33 & \textbf{86.26} & 0.008 & 96.57 & 81.33 & 0.009 & 95.83 & \textbf{97.44} & \textbf{0.03} & \textbf{92.07} & 99.20 \\
    Ensemble & \textbf{0.027} & \textbf{92.56} & 84.84 & \textbf{0.01} & \textbf{88.85} & 81.24 & \textbf{0.006} & \textbf{97.37} & 79.55 & \textbf{0.003} & \textbf{98.29} & 96.8 & 0.1 & 75.86 & \textbf{99.77} \\
    \midrule
    \textit{Nordic DSL} & \multicolumn{3}{c}{n=14,960} & \multicolumn{3}{c}{n=14,960} & \multicolumn{3}{c}{n=14,960} & \multicolumn{3}{c}{n=14,960} & \multicolumn{3}{c}{n=14,960} \\
    \midrule
    OpenLID-v3 & 0.007 & 96.04 & 83.22 & 0.012 & 94.10 & 93.56 & 0.017 & 91.98 & \textbf{96.87} & 0.002 & 99.07 & 93.33 & 0.039 & 92.72 & 99.44 \\ 
    OpenLID-v2 & 0.02 & 91.85 & 85.37 & 0.007 & 96.12 & 89.78 & 0.02 & 91.97 & 96.75 & 0.002 & 98.77 & 92.73 & 0.04 & 92.82 & 99.56 \\
    GlotLID & 0.01 & 94.58 & \textbf{94.50} & 0.006 & 97.09 & \textbf{93.77} & 0.006 & 97.20 & 96.47 & 0.004 & 98.16 & \textbf{93.69} & \textbf{0.025} & \textbf{95.16} & 99.36 \\
    Ensemble & \textbf{0.005} & \textbf{97.19} & 82.78 & \textbf{0.003} & \textbf{98.17} & 91.82 & \textbf{0.004} & \textbf{98.13} & 95.92 & \textbf{0.001} & \textbf{99.35} & 92.01 & 0.079 & 86.32 & \textbf{99.74} \\
    \midrule
    \textit{FLORES+} & \multicolumn{3}{c}{n=1,012} & \multicolumn{3}{c}{n=1,012} & \multicolumn{3}{c}{n=1,012} & \multicolumn{3}{c}{n=1,012} & \multicolumn{3}{c}{n=212,520} \\
    \midrule
    OpenLID-v3 & 1e-4 & 97.61 & 96.84 & 3e-5 & 99.31 & 99.21 & 7e-5 & 98.42 & \textbf{98.22} & \textbf{0} & \textbf{1.0} & 99.90 & 3e-3 & 99.99 & \textbf{1.0} \\ 
    OpenLID-v2 & 1e-4 & 96.84 & 96.84 & 2e-5 & 99.60 & 98.12 & 9e-5 & 98.12 & \textbf{98.22} & 0 & \textbf{1.0} & 99.90 & 0.004 & 99.99 & 99.99 \\
    GlotLID & 1e-4 & 97.10 & \textbf{99.41} & 1e-5 & 99.70 & \textbf{99.51} & \textbf{1e-5} & 99.67 & 97.53 & \textbf{0} & \textbf{1.0} & \textbf{1.0} & \textbf{0} & \textbf{1.0} & \textbf{1.0} \\
    Ensemble & \textbf{7e-5} & \textbf{98.39} & 96.74 & \textbf{6e-6} & \textbf{99.80} & 98.91 & \textbf{1e-5} & \textbf{99.70} & 97.23 & \textbf{0} & \textbf{1.0} & 99.90 & 1e-2 & 99.98 & \textbf{1.0} \\
    \midrule
    \textit{UDHR} & \multicolumn{3}{c}{n=62} & \multicolumn{3}{c}{n=61} & \multicolumn{3}{c}{n=58} & \multicolumn{3}{c}{n=61} & \multicolumn{3}{c}{n=27,515} \\
    \midrule
    OpenLID-v3 & \textbf{4e-5} & 98.39 & 98.39 & 4e-5 & 98.36 & \textbf{98.36} & 8e-5 & 96.61 & 98.28 & 7e-5 & 96.83 & \textbf{1.0} & 8e-3 & 99.99 & 99.98 \\ 
    OpenLID-v2 & 7e-5 & 96.88 & 1.0 & 7e-5 & 96.77 & 98.36 & 4e-5 & 98.28 & 98.28 & 1e-4 & 95.31 & 1.0 & 0.004 & 99.99 & 99.97 \\
    GlotLID & \textbf{4e-5} & \textbf{98.41} & \textbf{1.0} & 1e-4 & 93.75 & \textbf{98.36} & 4e-5 & 98.31 & \textbf{1.0} & 4e-5 & 98.39 & \textbf{1.0} & 0 & \textbf{1.0} & 99.98 \\
    Ensemble & \textbf{4e-5} & 98.39 & 98.39 & \textbf{0} & \textbf{1.0} & \textbf{98.36} & \textbf{0} & \textbf{1.0} & 98.28 & \textbf{0} & \textbf{1.0} & \textbf{1.0} & 8e-3 & 99.99 & \textbf{1.0} \\
    \bottomrule
\end{tabular}
}
\caption{Performance comparison across four benchmark datasets showing the false positive rate (FPR), precision (Prec.), and recall (Rec.) for Scandinavian and other languages. \textit{OpenLID-v3 }uses softmax threshold 0.5. \textit{Ensemble} represents top-1 agreement between \textit{OpenLID-v3} and \textit{GlotLID}. Best values per metric and language are in \textbf{bold}.}
\label{tab:sca-thresholds}
\end{table*}

\subsection{Scandinavian Languages}

Lastly, we focus on the North Germanic languages of Mainland Scandinavia, i.e. Norwegian Bokmål, Norwegian Nynorsk, Danish and Swedish. In addition to FLORES+ and UDHR, which cover all four languages, we experiment with two language-group-specific datasets:

First, we use \textbf{SLIDE} \citep{fedorova-etal-2025-multi}, a multilabel dataset based on clean instances from Universal Dependencies \cite{nivre-etal-2020-universal} (6,950 samples in total), containing many \langlabel{other} languages. Second, we use \textbf{Nordic DSL} \citep{haas-derczynski-2021-discriminating}, a single-label dataset containing  noisy sentences sourced from Wikipedia and featuring Faroese and Icelandic, in addition to the four aforementioned languages. 

We did not use the \textbf{FastSpell} dataset for the calculation of metrics, as upon manual examination, we noticed a large share of its Nynorsk samples to be valid in both Bokmål and Nynorsk. We re-annotated this subset and evaluated it separately.

Table~\ref{tab:sca-thresholds} summarizes the results. We find that GlotLID obtains the best recall, while OpenLID shows better false positive rate and precision. Precision and FPR are further improved by ensembling GlotLID and OpenLID-v3. OpenLID-v3 performs on par or slightly better than OpenLID-v2 on all benchmarks. Furthermore, the confusion matrices (available in the repository\punctfootnote{\url{https://github.com/ltgoslo/slide/tree/main/src/eval\_logs/vardial\_2026}}) show that the two Norwegian varieties, Bokmål and Nynorsk, are most easily confused. Appendix~\ref{sec:appendix-scandinavian} provides additional results: Table \ref{tab:sca-openlidversions} shows the effect of applying softmax threshold across all models and datasets, and Table \ref{tab:sca-ensemble-agreement} shows the effect of ensembling across datasets. Both tables report multilabel classification metrics.

\paragraph{HPLT-LID} The HPLT 3.0 manual inspection data \cite{oepen2025hplt30largescalemultilingual} only covers one Scandinavian language, namely Norwegian Bokmål. 402 documents were annotated as being correctly predicted Bokmål or not. We process this annotation in the same way as we did for BCMS annotation, excluding porn, unnatural language and artifacts. Documents with correctly identified language formed the dataset of documents guaranteed to be Bokmål, with 304 samples in total. 5 documents were annotated to be in a language other than Bokmål. We manually reviewed predictions of these 5 false positive Bokmål documents. Both OpenLID-v3 and GlotLID still predict them to be Bokmål. 3 of them are non-fluent Bokmål documents with many grammatical mistakes (articles of the wrong gender, articles and verbs omitted), as if written by foreign speakers, and 2 include code-switching, with Bokmål still being the predominant language. 

\paragraph{FastSpell.} A native speaker of Norwegian relabeled the FastSpell Nynorsk subset in a multilabel way, finding 40\% of samples to be (also) valid Bokmål. The annotator also reported that many word sequences were not full sentences and some texts were just menu choices from a web page. The evaluation of OpenLID-v3 and GlotLID on the resulting dataset is presented in Table~\ref{tab:nno}. While both models performed poorly in Bokmål detection, OpenLID-v3 performed slightly better for Nynorsk and produced fewer Danish and Swedish false positives, but more \langlabel{other} false positives than GlotLID.

\begin{table}
\resizebox{\linewidth}{!}{%
\begin{tabular}{lrrrrr}
    \toprule
    \textbf{Model} & \textbf{NB F1} & \textbf{NN F1} & \textbf{DA FP} & \textbf{SV FP} & \textbf{Other FP} \\
    \midrule
    OpenLID-v3 & 71.90 & \textbf{92.19} & \textbf{16} & \textbf{8} & 52 \\ 
    GlotLID & \textbf{79.07} & 91.14 & 27 & 19 & \textbf{16} \\
    \bottomrule
\end{tabular}
}
\caption{GlotLID (no threshold) and OpenLID-v3 (softmax threshold 0.5) on the re-annotated FastSpell Nynorsk data. F1 is for loose F1, FP is for false positives. 406 Nynorsk samples, 163 Bokmål samples.}
\label{tab:nno}
\end{table}

The manual analysis of the predictions showed that errors are often caused by named entities valid in any language using Latin script (``Georg Johannes Toft''), dates (``Levering 15. juli 2018'') and foreign words (``tomter i Parque natural cabo de gata''). These findings correspond to the error analysis from \citet{fedorova-etal-2025-multi} and further prove that shorter sequences, especially from the web, are the hardest for LID models.

In conclusion, we recommend OpenLID-v3 for general Scandinavian language identification if precision is prioritized, and GlotLID if recall is prioritized. For discriminating between Bokmål and Nynorsk, it is recommended to use a LID model trained specifically to separate these languages. 


\section{Conclusion}

We have evaluated the performance of three LID models (two OpenLID versions and GlotLID) on four multilingual benchmarks (two based on `clean' text and two based on web documents). Our new model, OpenLID-v3, performs on par or better in precision compared to its predecessors OpenLID-v2 and GlotLID, while the best precision is achieved by top-1 ensembling with GlotLID.
We have also shown that reliable evaluation of LID models including similar languages should be done on benchmarks specific to these languages, even if overall metrics on large multilingual benchmarks are high. Since web data may contain short texts valid in more than one language, there is a need for more multilabel training data and benchmarks.

\section*{Limitations}

The first limitation that we acknowledge is data mismatch between the evaluation and the intended use -- we would ideally evaluate on web text. However, large-scale web LID data of sufficient quality were not available at the time of conducting our experiment. We are aware of  the initiative by \citet{suarez2026commonlidreevaluatingstateoftheartlanguage}, but leave evaluation on their dataset for future work.

We have done our best to control for data contamination, where there was a known intersection of training and test data. However, this was not possible for tests on Nordic DSL, as this dataset, while consisting of Wikipedia data, was heavily preprocessed by its authors.

A common limitation of all models under evaluation is the unavailability of fully parallel datasets for all languages of interest, which might cause models to overfit to some concepts and named entities that are more common in certain languages.

\section*{Ethical Considerations}

All the new data annotations were done by the authors
voluntarily and without any monetary compensation.

We did not check any of OpenLID-v3's training data for inappropriate or biased content. We believe it does not do much harm, since the model is not generative. However, we can imagine some model predictions to be biased, particularly if used not as a classifier, but as a source of semantic representations.

A large amount of the current research is focused on collecting data for training large instruction-tuned generative models, capable of outputting grammatically and orthographically correct standard language, because of the high demand for such models in the public sector. This may result in loss of data written in low-resource varieties of mainstream languages, along with possibly valuable cultural knowledge. 

\section*{Acknowledgements}
This project was supported by the European Union Horizon Europe project no.\ 101070350 (HPLT). The computations were performed on resources provided by Sigma2 -- the National Infrastructure for High-Performance Computing and Data Storage in Norway.

We acknowledge the efforts of all the training data creators, including \citep{el2018arabic,ojha2019english,gongora-etal-2021-experiments,wong2015hong,alsarsour2018dart,abu2018shami,mdhaffar2017sentiment,meftouh2015machine,zaidan2011arabic,zahir2022iadd,goldhahn2012building,brown2012finding,bouamor2018madar,huidrom2021corpus,kashefi2018mizan,koehn2005europarl,neubig11kftt,tiedemann-2012-parallel,post-etal-2012-constructing,ziemski-etal-2016-unpc,rozis-skadins-2017-tilde,kunchukuttan-etal-2018-iit,agic-vulic-2019-jw300,espla-etal-2019-paracrawl,qi-etal-2018-pretrainemb,haddow2020pmindia,schwenk-etal-2019-wikimatrixv1,zhang-etal-2020-multiling-nmt,bojar-etal-2013-findings,bojar-etal-2014-findings,bojar-etal-2015-findings,bojar-etal-2016-findings,bojar-etal-2017-findings,bojar-etal-2018-findings,barrault-etal-2019-findings,barrault-etal-2020-findings,nllb2022,agic2014setimes,pilevar2011tep,mirzakhalov2021large,thoma2018wili,hasan-etal-2021-xl,winata2022nusax,lastrucci-etal-2023-preparing,Adelani2023MasakhaNEWS,ogundepo2023afriqa}.

\bibliography{anthology.sm,custom,openlid_dataset_citations}

\appendix

\section{Multilingual Benchmarks}
\label{app:hplt30}

In all evaluations on all datasets, metrics were only calculated on subset of languages, present in all models being compared. For GlotLID, we always used its the most recent version at the moment of writing, GlotLID-v3.

Table~\ref{tab:hplt30} shows statistics of the HPLT-LID dataset. Only those samples were accepted, which were not unnatural language, porn, web artifacts or incorrect LID. One must notice that, while this dataset contains texts from the web, it can not be used as a benchmark of noisy web data, because the texts were heavily cleaned and preprocessed before their annotation was performed. We believe however that HPLT-LID may be valuable for the community, because there exist not many LID benchmarks for some languages presented there..

Table \ref{tab:hplt-lid} presents evaluation of all models (except OpenLID-v2, because the dataset was created with its help) under consideration on HPLT-LID. This is just another proof that there were no regression in performance of OpenLID-v3 compared to OpenLID-v2.

\begin{table}[]
\centering
\begin{tabular}{lclc}
    \toprule
    Language & \# docs & Language & \# docs \\
    \midrule
    \langlabel{ast\_Latn} & 139 &	\langlabel{hrv\_Latn} & 188 \\
\langlabel{bos\_Latn} & 49 &	\langlabel{ita\_Latn} & 171 \\
\langlabel{cat\_Latn} & 196 &	\langlabel{jpn\_Jpan} & 87 \\
\langlabel{cmn\_Hans} & 183 &	\langlabel{nob\_Latn} & 304 \\
\langlabel{ces\_Latn} & 183 &	\langlabel{pes\_Arab} & 209 \\
\langlabel{deu\_Latn} & 179 &	\langlabel{por\_Latn} & 111 \\
\langlabel{ell\_Grek} & 613 &	\langlabel{rus\_Cyrl} & 362 \\
\langlabel{eng\_Latn} & 126 &	\langlabel{spa\_Latn} & 300 \\
\langlabel{fra\_Latn} & 139 &	\langlabel{slk\_Latn} & 127 \\
\langlabel{fin\_Latn} & 335 &	\langlabel{srp\_Cyrl} & 15 \\
\langlabel{glg\_Latn} & 182 &	\langlabel{yor\_Latn} & 168 \\
\langlabel{hin\_Deva} & 136 &	total & 4,502 \\
    \bottomrule
\end{tabular}%
\caption{HPLT 3.0 manual inspection dataset statistics.}
\label{tab:hplt30}
\end{table}

\begin{table}[]
    \centering
    \resizebox{\linewidth}{!}{%
    \begin{tabular}{l|rrrr}
    \toprule
    Model  & FPR & F1 & Precision & Recall \\ 
    \midrule
    OpenLID-v3, 0.5 & 4e-5 & \textbf{99.84} & 99.91 & 99.78 \\ 
    OpenLID-v3 & 6e-5 & \textbf{99.84} & 99.85 & \textbf{99.84} \\ 
    GlotLID, 0.5 & 4e-5 & 99.51 & 99.89 & 99.16 \\ 
    GlotLID  & 4e-5 & 99.55 & 99.89 & 99.24 \\ 
    ensemble & \textbf{1e-5} & 99.53 & \textbf{99.96} & 99.14 \\ 
    \bottomrule
    \end{tabular}
    }
    \caption{Evaluation on HPLT-LID (n=3,873), with and without softmax threshold.}
    \label{tab:hplt-lid}
\end{table}

\begin{table}
    \centering
    \resizebox{\linewidth}{!}{%
    \begin{tabular}{l|rrrr}
    \toprule
       Model  & FPR & F1 & Precision & Recall \\
    \midrule
    \textit{FastSpell} & \multicolumn{4}{c}{n=6,809} \\
    \midrule
    OpenLID-v3, 0.5 & 2.5e-3          & 91.08          & 95.72          & 89.56 \\
    OpenLID-v3      & 2.9e-3          & 91.22          & 95.23          & 90.32 \\
    OpenLID-v2, 0.5 & 3.9e-3          & 90.04          & 94.55          & 89.07 \\
    OpenLID-v2      & 4.2e-3          & 90.37          & 94.05          & 90.11 \\
    GlotLID, 0.5    & 3.0e-3          & 91.32          & 95.19          & 90.24  \\
    GlotLID         & 3.5e-3          & \textbf{91.39} & 94.63          & \textbf{91.00} \\
    ensemble        & \textbf{1.5e-3} & 90.92          & \textbf{97.11} & 87.78 \\
    \midrule
    \textit{FLORES+} & \multicolumn{4}{c}{n=155,848} \\
    \midrule
    OpenLID-v3, 0.5   & 7.6e-5          & 97.72          & 97.86          & 97.71 \\
    OpenLID-v3        & 8.6e-5          & 97.74          & 97.70          & 97.89 \\
    OpenLID-v2, 0.5   & 8.9e-5          & 98.09          & 98.62          & 97.98 \\
    OpenLID-v2        & 1.0e-4          & 98.11          & 98.46          & \textbf{98.16} \\
    GlotLID, 0.5      & 3.4e-5          & 98.34          & \textbf{99.52} & 97.95 \\
    GlotLID           & 3.8e-5          & \textbf{98.38} & 99.46          & 98.07 \\
    ensemble          & \textbf{2.6e-5} & 97.01          & 98.33          & 96.5 \\    
    \midrule
    \textit{UDHR} & \multicolumn{4}{c}{n=10,283} \\
    \midrule
    OpenLID-v3, 0.5      & 1.5e-4          & 96.14          & 97.13          & 96.39 \\
    OpenLID-v3           & 1.8e-4          & 96.21          & 96.87          & 96.22 \\
    OpenLID-v2, 0.5      & 2.5e-4          & 96.27          & 97.26          & 96.73 \\
    OpenLID-v2           & 3.0e-4          & 96.21          & 96.93          & \textbf{96.91} \\
    GlotLID, 0.5         & 6.7e-5          & \textbf{97.39} & \textbf{98.96} & 96.76 \\
    GlotLID              & 8.7e-5          & 97.37          & 98.78          & 96.87 \\
    ensemble             & \textbf{4.5e-5} & 95.59          & 98.07          & 94.58 \\ 
    \bottomrule
    \end{tabular}
    }
    \caption{Multilingual benchmarks. F1, precision and recall in \%. `0.5' stands for thresholding softmax scores at 0.5. Ensembling experiments are done with thresholded OpenLID-v3 and GlotLID without thresholding. FLORES+ refers to the devtest split, and FastSpell refers to our version that excludes Nynorsk.}
    \label{tab:many}
\end{table}

\section{Additional Data Sources}
\label{sec:appendix-data-sources}

Table \ref{tab:appendix-data-sources} shows the difference between OpenLID-v2 and OpenLID-v3 in terms of training data sources.  

\paragraph{Languages outside OpenLID-v2 class inventory.} We discussed adding natural language \langlabel{other} class (languages present in glotlid-corpus other than our 190+) for collecting `garbage' documents and out-of-train-set languages. However, after looking closer into the data, just sampling from about 1,900 those languages and labeling them all as "other" did not seem a good idea. About 1,500 of those language were low-resource ones where only Bibles were available, reported as non-reliable data by GlotLID's authors\footnote{\url{https://github.com/cisnlp/GlotLID/blob/main/sources.md}}.

After removing non-reliable sources, 366 languages were left. The smallest language in OpenLID-v2 dataset is Yiddish with 923 samples only. Out of 366, we have 150 languages with more samples than that. Some of them were quite large like Low German (\langlabel{nds\_Latn}) with 117,676 samples (many languages in OpenLID-v2 dataset have less), national like Romansh (\langlabel{roh\_Latn}) or have more than a million speakers like Chechnyan (\langlabel{che\_Cyrl}). For the future OpenLID improvement, we recommend extending the model with these languages as separate labels instead of pushing them all into "other". 

Not all languages with many samples are expected to be common in crawls: some of them are not alive like Classical Chinese (\langlabel{lzh\_Hani}) or constructed like Interlingua (\langlabel{ina\_Latn}). Since OpenLID already featured Esperanto and Latin, we believe other such language are also worth being edited.

Also we found more languages present in both Cyrillic and Latin script, like Serbian: Tatar, for example.

\paragraph{Training data for \langlabel{zxx\_Zxxx}.} \langlabel{und\_*} are random sequences generated from different scripts.\punctfootnote{\url{https://github.com/cisnlp/GlotLID/blob/main/assets/train/gen_und.py}} \langlabel{zxx\_*} are non-linguistic noise  in different scripts collected from the web.\punctfootnote{\url{https://huggingface.co/datasets/cis-lmu/glotlid-corpus/tree/main/zxx}} (We experimented with adding them as separated classes first; while our validation data had no such class, some languages outside our class inventory were classified as one of this classes; the choice between the two was random according to manual inspection.)

\begin{table*}[]
\resizebox{\linewidth}{!}{%
\begin{tabular}{l|ll}
    \toprule
    Language & Sources & Justification \\
    \midrule
    \langlabel{ast\_Latn} & commonvoice, leipzigwiki, tatoeba & high confusion with \langlabel{other} \\
    \langlabel{bam\_Latn} & koumankanmtdyufr\tablefootnote{\url{https://huggingface.co/datasets/uvci/Koumankan\_mt\_dyu\_fr}}, ud, all OpenLID-v2 dyu data & high confusion with \langlabel{dyu\_Latn} \\
    \langlabel{ben\_Beng} & commonvoice, GlotStoryBook, leipzigwiki, tatoeba, ud & small in HPLT 3.0 \\
    \langlabel{bod\_Tibt} & GlotStoryBook, tatoeba, the newest Wikipedia & high confusion with \langlabel{dzo\_Tibt} \\
    \langlabel{cat\_Latn} & commonvoice, GlotStoryBook, leipzigwiki, tatoeba, ud & some confusion with \langlabel{oci\_Latn} \\
    \langlabel{dan\_Latn} & commonvoice, GlotStoryBook, leipzigwiki, tatoeba, ud & high confusion with \langlabel{nob\_Latn} \\
    \langlabel{dzo\_Tibt} & the newest Wikipedia &  high confusion with \langlabel{bod\_Tibt} \\
    \langlabel{frp\_Latn} & HFWikipedia & missed in OpenLID-v2, high confusion with \langlabel{lij\_Latn} \\
    \langlabel{fuv\_Latn} & tatoeba & high confusion with \langlabel{other} \\
    \langlabel{gug\_Latn} & commonvoice, leipzigwiki, tatoeba, ud & poor in HPLT 3.0 \\
    \langlabel{heb\_Hebr} & leipzigwiki, tatoeba, ud & minor confusion with underrepresented Yiddish\\
    \langlabel{kan\_Knda} & GlotStoryBook, leipzigwiki, tatoeba & small in HPLT 3.0 \\
    \langlabel{kin\_Latn} & afriqa, commonvoice, GlotStoryBook, tatoeba & high confusion with \langlabel{run\_Latn} \\
    \langlabel{lat\_Latn} & tatoeba, ud, GlotStoryBook, leipzigwiki & missed in OpenLID-v2 \\
    \langlabel{oci\_Latn} & removing OpenLID's `pilar' & high confusion with \langlabel{lij\_Latn} \\
    \langlabel{run\_Latn} & masakhanews, tatoeba, the newest Wikipedia & high confusion with \langlabel{kin\_Latn} \\
    \langlabel{rus\_Cyrl} & commonvoice, GlotStoryBook, leipzigwiki, tatoeba, ud & high confusion with \langlabel{other} \\
    \langlabel{som\_Latn} & GlotStoryBook, leipzigwiki, masakhanews, tatoeba & high confusion with \langlabel{other} \\
    \langlabel{srp\_Latn} & ud, setimes, tatoeba & missed in OpenLID-v2, high confusion with Croatian and Bosnian \\
    \langlabel{ssw\_Latn}  & GlotStoryBook, Vukuzenzele, the newest Wikipedia & high confusion with \langlabel{other} \\
    \langlabel{sun\_Latn} & leipzigwiki, nusa, tatoeba & high confusion with \langlabel{other} \\
    \langlabel{tam\_Taml} & commonvoice, GlotStoryBook, leipzigwiki, tatoeba, ud & small in HPLT 3.0 \\
    \langlabel{zxx\_Zxxx} & \langlabel{zxx\_*}, \langlabel{und\_*} & missed in OpenLID-v2 \\
    \bottomrule
\end{tabular}%
}
\caption{OpenLID-v3 added/removed training data, compared to OpenLID-v2; added data were taken from GlotLID-corpus, if another not specified.}
\label{tab:appendix-data-sources}
\end{table*}

\section{Bosnian, Croatian, Montenegrin and Serbian}
\label{app:hbs}

\begin{table}[]
    \centering
    \begin{tabular}{crrr}
 \toprule
\textit{(gold/pred)} & \texttt{bos} & \texttt{hrv} & \texttt{srp} \\ \midrule
 \texttt{bos}  & \textbf{1,922} &  571 & - \\
 \texttt{hrv}  &  534 & \textbf{1,833} & - \\
 \texttt{srp}  & \textbf{6,129} &  468 & 3,694 \\
 \bottomrule
    \end{tabular}
    \caption{GlotLID evaluation on SETimes; average document length by character count, across correctly and incorrectly labeled data.}
    \label{tab:doclen}
\end{table}

\paragraph{Other benchmarks} We are aware of BCS news dataset (SETimes) \citep{clarin_setimes}, comprising roughly 9,000 parallel documents written in each of the languages, from a now-inactive news site that published articles in the languages of South-Eastern Europe. However, 
SETimes data were present among both OpenLID and GlotLID's training data, but their splitting into sentences seem to be done in different ways. While we succeeded in deduplicating the test data from GloLID's training instances, we scored suspiciously high with OpenLID (more than 90\% F1), which was never observed on other benchmarks. This is a clear sign that our deduplication has not worked. Thus, we only report GlotLID results here and leave proper evaluation on SETimes for future work. It is an important benchmark, because it comprises multi-sentence documents; average document length for each combination of gold and assigned label is given in Table \ref{tab:doclen}. Generally, the common wisdom that longer documents lead to better classification holds here as well, mainly because longer documents give more opportunities for clear grammatical markers and discriminators to arise. Short sentences are much more likely to be mislabeled, across all models and datasets. Note that Serbian (\texttt{srp}, bottom row) doesn't follow the pattern of longer documents being more likely to be correctly labeled. In addition to performance on Latin-script Serbian generally being lower, this discrepancy is likely due to common lexical overlap between Bosnian and Serbian, exacerbated by document length, and the model seemingly giving preference to the \texttt{bos} label, discussed further in this section.

We also acknowledge the existence of the COPA \citep{rupnik-etal-2023-benchic} and Heritage BCS \citep{romic2024heritage} datasets, but, as the Serbian portion of COPA uses the standard Cyrillic script only, it was inappropriate for our evaluation; and Heritage BCS data comprise transcripts of bilingual speakers' conversations, there is too much second-language interference in the documents for meaningful evaluation on LID task.

\paragraph{Montenegrin GlotLID predictions} Interestingly, there are a handful of mislabeled documents that GlotLID labels as Montenegrin (\texttt{cnr}) or Chakavian (\texttt{ckm}, a dialect of Croatian). Both the first\footnote{\textit{ ``Dijagnoza je točna, nisam i siguran da je i terapija, nisam siguran da je i terapija ispravna.''}} and the second\footnote{\textit{ ``Bijele noći, pa normalno da su bijele noći smetale, sve što je bijelo smetalo je, u mraku se puno bolje rade ovakvi poslovi.''}} feature nonstandard syntax and repetitive \textit{i} conjunctions and adjectives; both are, respectively, typical of Chakavian and Montenegrin poetry, which feature disproportionately in training data. 

\paragraph{HPLT-LID reannotation details} 

201 samples were predicted by OpenLID-v2 to be Serbian Cyrillic, out of which 151 were pre-annotated to be false positives in the corresponding repository, but turned out to be missing annotation rather then false. 21 samples were detected to be translationese from Serbian Cyrillic and annotated with a new tag \langlabel{tra\_Zxxx}; 3 samples were found to be only valid in Montenegrin and annotated with \langlabel{cnr\_Latn}.

\section{Romance Languages of Italy and France}
\label{sec:appendix-lij-oci-frp}

Table \ref{tab:oci_frp} lists the Francoprovençal and Occitan translations of the UDHR and their corresponding URLs. 

\begin{table*}[]
    \centering
    \begin{tabular}{l|l}
    \toprule
       Dialect  & URL \\
    \midrule
    Occitan Auvergnat & \url{http://efele.net/udhr/d/udhr_auv.txt} \\
    Occitan Languedocien & \url{http://efele.net/udhr/d/udhr_lnc.txt} \\
    Occitan Provençal & \url{http://efele.net/udhr/d/udhr_prv.txt} \\
    Francoprovençal Fribourg & \url{http://efele.net/udhr/d/udhr_oci_1.txt} \\
    Francoprovençal Savoie & \url{http://efele.net/udhr/d/udhr_oci_2.txt} \\
    Francoprovençal Vaud  & \url{http://efele.net/udhr/d/udhr_oci_3.txt} \\
    Francoprovençal Valais & \url{http://efele.net/udhr/d/udhr_oci_4.txt} \\
    \bottomrule
    \end{tabular}
    \caption{Occitan and Francoprovençal UDHR translations and their URLs.}
    \label{tab:oci_frp}
\end{table*}

\section{Scandinavian Languages}
\label{sec:appendix-scandinavian}

Table \ref{tab:sca-ensemble-agreement} presents ensemble results combining GlotLID (no threshold) and OpenLID-v3 (softmax threshold 0.5).
Table \ref{tab:sca-openlidversions} shows performance comparison for different versions of OpenLID with GlotLID on the SLIDE, Nordic DSL, FLORES+ devtest, and UDHR test sets with and without thresholding.

\begin{table*}[]
\centering
\begin{tabular}{lrrrrrrr}
    \toprule
    \textbf{Dataset} & \textbf{Loose Acc.} & \textbf{Exact Acc.} & \textbf{NB F1} & \textbf{DA F1} & \textbf{NN F1} & \textbf{SV F1} & \textbf{Other F1} \\
    \midrule
    SLIDE & \textit{94.62} & \textit{92.30} & 92.52 & \textit{90.53} & 91.52 & \textit{98.21} & 91.70 \\ 
    Nordic DSL & 93.66 & 93.66 & \textit{89.41} & \textit{94.89} & \textbf{97.01} & 95.54 & 92.54 \\
    FLORES+ & 99.97 & 99.97 & \textit{97.56} & \textit{99.35} & \textit{98.45} & 99.95 & 99.99 \\
    UDHR & \textbf{99.99} & \textbf{99.99} & 98.39 & \textbf{99.17} & \textit{99.13} & \textbf{100.00} & \textbf{100.00} \\
    \bottomrule
\end{tabular}%
\caption{Ensembling GlotLID (no threshold) and OpenLID-v3 (softmax threshold 0.5) on Scandinavian languages. Bold indicates improvement over both individual models. Italic indicates improvement over OpenLID-v3 only.}
\label{tab:sca-ensemble-agreement}
\end{table*}

\begin{table*}[]
\centering
\resizebox{\linewidth}{!}{%
\begin{tabular}{ll|rr|rrrrr}
\toprule
\textbf{Thres.} & \textbf{Model} & \textbf{Loose Acc.} & \textbf{Exact Acc.} & \textbf{NB F1} & \textbf{DA F1} & \textbf{NN F1} & \textbf{SV F1} & \textbf{Other F1} \\
\midrule
\multicolumn{9}{l}{\textit{SLIDE}} \\
\midrule
\multirow{4}{*}{No} & OpenLID-v3 & 94.46 & 90.55 & 94.33 & 89.27 & 93.61 & 96.30 & 93.13 \\ 
& OpenLID-v2 & 94.39 & 90.36 & 94.07 & 89.60 & 94.15 & 95.89 & 93.11 \\
& OpenLID-v1 & 93.61 & 89.77 & 93.91 & 87.56 & 93.03 & 95.51 & 91.42 \\
& GlotLID & \textbf{97.20} & \textbf{93.40} & \textbf{95.83} & \textbf{93.37} & \textbf{94.84} & \textbf{98.23} & \textbf{98.05} \\
[0.5ex]
\multirow{4}{*}{0.5} & OpenLID-v3 & 95.55 & 91.87 & 94.25 & 91.60 & 93.52 & 97.62 & 95.16 \\ 
& OpenLID-v2 & 95.63 & 91.83 & 94.16 & 91.58 & 94.12 & 97.38 & 95.29 \\
& OpenLID-v1 & 94.81 & 91.14 & 94.00 & 89.26 & 93.36 & 96.84 & 93.76 \\
& GlotLID & \textbf{97.09} & \textbf{93.45} & \textbf{95.56} & \textbf{93.40} & \textbf{94.52} & \textbf{98.22} & \textbf{97.73} \\
[0.5ex]
& n samples & 6,950 & 6,950 & 2,098 & 677 & 1,628 & 1,250 & 1,745 \\
\midrule
\multicolumn{9}{l}{\textit{Nordic DSL (from 50k split)}} \\
\midrule
\multirow{4}{*}{No} & OpenLID-v3 & 94.71 & 94.71 & 89.29 & 93.63 & 94.24 & \textbf{96.06} & \textbf{97.36} \\ 
& OpenLID-v2 & 94.32 & 94.32 & 88.63 & 92.89 & 93.79 & 95.67 & 97.30 \\
& OpenLID-v1 & 94.17 & 94.17 & 88.73 & 93.64 & 92.78 & 95.37 & 97.15 \\
& GlotLID & \textbf{96.19} & \textbf{96.19} & \textbf{94.54} & \textbf{95.40} & \textbf{96.84} & 95.87 & 97.21 \\
[0.5ex]
\multirow{4}{*}{0.5} & OpenLID-v3 & 94.30 & 94.30 & 89.17 & 93.83 & 94.36 & \textbf{96.11} & 95.96 \\ 
& OpenLID-v2 & 93.96 & 93.96 & 88.49 & 92.84 & 94.30 & 95.65 & 96.07 \\
& OpenLID-v1 & 93.88 & 93.88 & 88.47 & 93.65 & 93.16 & 95.41 & 96.13 \\
& GlotLID & \textbf{96.10} & \textbf{96.10} & \textbf{94.48} & \textbf{95.51} & \textbf{96.85} & 95.96 & \textbf{96.86} \\
[0.5ex]
& n samples & 74,800 & 74,800 & 14,960 & 14,960 & 14,960 & 14,960 & 14,960 \\
\midrule
\multicolumn{9}{l}{\textit{FLORES+ devtest}} \\
\midrule
\multirow{4}{*}{No} & OpenLID-v3 & 99.97 & 99.97 & 97.53 & 99.11 & 98.22 & 99.90 & 100.00 \\ 
& OpenLID-v2 & 99.97 & 99.97 & 97.19 & 98.91 & 98.08 & 99.90 & \textbf{100.00} \\
& OpenLID-v1 & 99.97 & 99.97 & 97.63 & 98.86 & 98.57 & 99.75 & \textbf{100.00} \\
& GlotLID & \textbf{99.98} & \textbf{99.98} & \textbf{98.24} & \textbf{99.60} & \textbf{98.60} & \textbf{100.00} & \textbf{100.00} \\
[0.5ex]
\multirow{4}{*}{0.5} & OpenLID-v3 & 99.97 & 99.97 & 97.22 & 99.26 & 98.32 & 99.95 & \textbf{100.00} \\ 
& OpenLID-v2 & 99.97 & 99.97 & 96.84 & 98.86 & 98.17 & 99.95 & \textbf{100.00} \\
& OpenLID-v1 & 99.97 & 99.97 & 97.53 & 98.86 & \textbf{98.62} & 99.85 & \textbf{100.00} \\
& GlotLID & \textbf{99.98} & \textbf{99.98} & \textbf{98.19} & \textbf{99.55} & 98.60 & \textbf{100.00} & \textbf{100.00} \\
[0.5ex]
& n samples & 216,568 & 216,568 & 1,012 & 1,012 & 1,012 & 1,012 & 212,520 \\
\midrule
\multicolumn{9}{l}{\textit{UDHR}} \\
\midrule
\multirow{4}{*}{No} & OpenLID-v3 & 99.75 & 99.75 & 93.85 & \textbf{97.44} & 82.19 & 78.21 & 99.88 \\ 
& OpenLID-v2 & 99.74 & 99.74 & 96.12 & 85.71 & 95.80 & 73.49 & 99.87 \\
& OpenLID-v1 & 99.79 & 99.79 & 83.22 & 95.31 & 91.20 & 88.41 & 99.90 \\
& GlotLID & \textbf{99.97} & \textbf{99.97} & \textbf{99.20} & 96.00 & \textbf{99.15} & \textbf{99.19} & \textbf{99.99} \\
[0.5ex]
\multirow{4}{*}{0.5} & OpenLID-v3 & 99.97 & 99.97 & 98.39 & 98.36 & 97.44 & 98.39 & 99.99 \\ 
& OpenLID-v2 & 99.97 & 99.97 & 98.41 & 97.56 & 98.28 & 97.60 & 99.99 \\
& OpenLID-v1 & 99.97 & 99.97 & 98.41 & \textbf{99.19} & 95.00 & 99.19 & 99.99 \\
& GlotLID & \textbf{99.99} & \textbf{99.99} & \textbf{99.20} & 98.36 & \textbf{100.00} & \textbf{100.00} & \textbf{100.00} \\
[0.5ex]
& n samples & 27,757 & 27,757 & 62 & 61 & 58 & 61 & 27,515 \\
\bottomrule
\end{tabular}%
}
\caption{Comprehensive comparison of language identification models across all test datasets with and without softmax thresholding. Bold values indicate best performance for each metric within each dataset/threshold combination.}
\label{tab:sca-openlidversions}
\end{table*}

\section{Hierarchical Models}
\label{app:cascade}

Hierarchical LID systems decompose the classification taks into a series of decision steps, typically progressing from broad language groups to the individual languages or variants \citep{jauhiainen2019automatic,agarwal-etal-2023-limit}. The
motivation for these approaches is that closely related languages are naturally more challenging to discriminate for LID models \citep{goutte-etal-2014-nrc}.

We trained hierarchical fastText-based models and evaluated their effect on the following language groups: Scandinavian languages (Danish, Faroese, Icelandic, Norwegian Bokm{\aa}l, Norwegian Nynorsk, and Swedish), South Slavic languages (Bosnian, Croatian, Serbian in Latin script), along with varieties of Arabic (Egyptian, Levantine, Mesopotamian, Moroccan, Najdi, Standard Arabic, Ta'izzi-Adeni, and Tunisian), and Persian (Dari and Iranian Persian). For each group, we train a specialized language classifier on the subset of the OpenLID-v3 corpus composed of these languages. During inference, every time the base model predicts a language from one of the four groups, we replace
the base model prediction with the prediction of the specialized model.

The specialized models follow the same architecture and training procedure as OpenLID-v3. For each language group, we use the corresponding language-specific subset of the OpenLID-v3 training data.

We have not found a significant difference in the performance of OpenLID-v3 and the hierarchical models in any of the related language groups when evaluating on FLORES+ or UDHR. We hypothesize that the capacity of the large LID model is not saturated yet and therefore there is nothing to be gained by using a subset of the training data. Instead, effort should be put into securing larger and higher-quality annotated data for these language groups.

\section{List of Supported Languages}
\label{app:openlidv3-languages}
Table \ref{tab:openlidv3-langs} shows the list of language labels supported by OpenLID-v3.

\begin{table*}[]
\centering
\resizebox{\linewidth}{!}{%
\begin{tabular}{lp{3.1cm}|lp{3.1cm}|lp{3.1cm}|lp{3.1cm}}
    \toprule
    Lang. code & Name & Lang. code & Name & Lang. code & Name & Lang. code & Name \\
    \midrule
    \langlabel{ace\_Arab} & Achinese &	\langlabel{fon\_Latn} & Fon &	\langlabel{lij\_Latn} & Ligurian &	\langlabel{slk\_Latn} & Slovak \\
\langlabel{ace\_Latn} & Achinese &	\langlabel{fra\_Latn} & French &	\langlabel{lim\_Latn} & Limburgan (Limburger, Limburgish) &	\langlabel{slv\_Latn} & Slovenian \\
\langlabel{afr\_Latn} & Afrikaans &	\langlabel{frp\_Latn} & Arpitan (Francoprovençal) &	\langlabel{lin\_Latn} & Lingala &	\langlabel{smo\_Latn} & Samoan \\
\langlabel{als\_Latn} & Tosk Albanian &	\langlabel{fur\_Latn} & Friulian &	\langlabel{lit\_Latn} & Lithuanian &	\langlabel{sna\_Latn} & Shona \\
\langlabel{amh\_Ethi} & Amharic &	\langlabel{fuv\_Latn} & Nigerian Fulfulde &	\langlabel{lmo\_Latn} & Lombard &	\langlabel{snd\_Arab} & Sindhi \\
\langlabel{ara\_Arab} & Arabic &	\langlabel{gaz\_Latn} & West Central Oromo &	\langlabel{ltg\_Latn} & Latgalian &	\langlabel{som\_Latn} & Somali \\
\langlabel{asm\_Beng} & Assamese &	\langlabel{gla\_Latn} & Scottish Gaelic &	\langlabel{ltz\_Latn} & Luxembourgish (Letzeburgesch) &	\langlabel{sot\_Latn} & Southern Sotho \\
\langlabel{ast\_Latn} & Asturian (Asturleonese, Bable, Leonese) &	\langlabel{gle\_Latn} & Irish &	\langlabel{lua\_Latn} & Luba-Lulua &	\langlabel{spa\_Latn} & Spanish (Castilian) \\
\langlabel{awa\_Deva} & Awadhi &	\langlabel{glg\_Latn} & Galician &	\langlabel{lug\_Latn} & Ganda &	\langlabel{srd\_Latn} & Sardinian \\
\langlabel{ayr\_Latn} & Central Aymara &	\langlabel{gug\_Latn} & Paraguayan Guaraní &	\langlabel{luo\_Latn} & Luo (Kenya and Tanzania) (Dholuo) &	\langlabel{srp\_Cyrl} & Serbian \\
\langlabel{azb\_Arab} & South Azerbaijani &	\langlabel{guj\_Gujr} & Gujarati &	\langlabel{lus\_Latn} & Lushai &	\langlabel{srp\_Latn} & Serbian \\
\langlabel{azj\_Latn} & North Azerbaijani &	\langlabel{hat\_Latn} & Haitian (Creole, Haitian Creole) &	\langlabel{lvs\_Latn} & Standard Latvian &	\langlabel{ssw\_Latn} & Swati \\
\langlabel{bak\_Cyrl} & Bashkir &	\langlabel{hau\_Latn} & Hausa &	\langlabel{mag\_Deva} & Magahi &	\langlabel{sun\_Latn} & Sundanese \\
\langlabel{bam\_Latn} & Bambara &	\langlabel{heb\_Hebr} & Hebrew &	\langlabel{mai\_Deva} & Maithili &	\langlabel{swe\_Latn} & Swedish \\
\langlabel{ban\_Latn} & Balinese &	\langlabel{hin\_Deva} & Hindi &	\langlabel{mal\_Mlym} & Malayalam &	\langlabel{swh\_Latn} & Swahili (individual language) (Kiswahili) \\
\langlabel{bel\_Cyrl} & Belarusian &	\langlabel{hne\_Deva} & Chhattisgarhi &	\langlabel{mar\_Deva} & Marathi &	\langlabel{szl\_Latn} & Silesian \\
\langlabel{bem\_Latn} & Bemba (Zambia) &	\langlabel{hrv\_Latn} & Croatian &	\langlabel{min\_Latn} & Minangkabau &	\langlabel{tam\_Taml} & Tamil \\
\langlabel{ben\_Beng} & Bengali &	\langlabel{hun\_Latn} & Hungarian &	\langlabel{mkd\_Cyrl} & Macedonian &	\langlabel{taq\_Latn} & Tamasheq \\
\langlabel{bho\_Deva} & Bhojpuri &	\langlabel{hye\_Armn} & Armenian &	\langlabel{mlt\_Latn} & Maltese &	\langlabel{taq\_Tfng} & Tamasheq \\
\langlabel{bjn\_Arab} & Banjar &	\langlabel{ibo\_Latn} & Igbo &	\langlabel{mni\_Beng} & Manipuri &	\langlabel{tat\_Cyrl} & Tatar \\
\langlabel{bjn\_Latn} & Banjar &	\langlabel{ilo\_Latn} & Iloko &	\langlabel{mos\_Latn} & Mossi &	\langlabel{tel\_Telu} & Telugu \\
\langlabel{bod\_Tibt} & Tibetan &	\langlabel{ind\_Latn} & Indonesian &	\langlabel{mri\_Latn} & Maori &	\langlabel{tgk\_Cyrl} & Tajik \\
\langlabel{bos\_Latn} & Bosnian &	\langlabel{isl\_Latn} & Icelandic &	\langlabel{mya\_Mymr} & Burmese &	\langlabel{tha\_Thai} & Thai \\
\langlabel{bug\_Latn} & Buginese &	\langlabel{ita\_Latn} & Italian &	\langlabel{nld\_Latn} & Dutch (Flemish) &	\langlabel{tir\_Ethi} & Tigrinya \\
\langlabel{bul\_Cyrl} & Bulgarian &	\langlabel{jav\_Latn} & Javanese &	\langlabel{nno\_Latn} & Norwegian Nynorsk &	\langlabel{tpi\_Latn} & Tok Pisin \\
\langlabel{cat\_Latn} & Catalan (Valencian) &	\langlabel{jpn\_Jpan} & Japanese &	\langlabel{nob\_Latn} & Norwegian Bokmål &	\langlabel{tsn\_Latn} & Tswana \\
\langlabel{ceb\_Latn} & Cebuano &	\langlabel{kab\_Latn} & Kabyle &	\langlabel{npi\_Deva} & Nepali (individual language) &	\langlabel{tso\_Latn} & Tsonga \\
\langlabel{ces\_Latn} & Czech &	\langlabel{kac\_Latn} & Kachin (Jingpho) &	\langlabel{nso\_Latn} & Pedi (Northern Sotho, Sepedi) &	\langlabel{tuk\_Latn} & Turkmen \\
\langlabel{cjk\_Latn} & Chokwe &	\langlabel{kam\_Latn} & Kamba (Kenya) &	\langlabel{nus\_Latn} & Nuer &	\langlabel{tum\_Latn} & Tumbuka \\
\langlabel{ckb\_Arab} & Central Kurdish &	\langlabel{kan\_Knda} & Kannada &	\langlabel{nya\_Latn} & Chichewa (Chewa, Nyanja) &	\langlabel{tur\_Latn} & Turkish \\
\langlabel{cmn\_Hans} & Mandarin Chinese &	\langlabel{kas\_Arab} & Kashmiri &	\langlabel{oci\_Latn} & Occitan (post 1500) &	\langlabel{twi\_Latn} & Twi \\
\langlabel{cmn\_Hant} & Mandarin Chinese &	\langlabel{kas\_Deva} & Kashmiri &	\langlabel{ory\_Orya} & Odia (Oriya (individual language)) &	\langlabel{uig\_Arab} & Uighur (Uyghur) \\
\langlabel{crh\_Latn} & Crimean Tatar (Crimean Turkish) &	\langlabel{kat\_Geor} & Georgian &	\langlabel{pag\_Latn} & Pangasinan &	\langlabel{ukr\_Cyrl} & Ukrainian \\
\langlabel{cym\_Latn} & Welsh &	\langlabel{kaz\_Cyrl} & Kazakh &	\langlabel{pan\_Guru} & Panjabi (Punjabi) &	\langlabel{umb\_Latn} & Umbundu \\
\langlabel{dan\_Latn} & Danish &	\langlabel{kbp\_Latn} & Kabiyè &	\langlabel{pap\_Latn} & Papiamento &	\langlabel{urd\_Arab} & Urdu \\
\langlabel{deu\_Latn} & German &	\langlabel{kea\_Latn} & Kabuverdianu &	\langlabel{pbt\_Arab} & Southern Pashto &	\langlabel{uzn\_Latn} & Northern Uzbek \\
\langlabel{dik\_Latn} & Southwestern Dinka &	\langlabel{khk\_Cyrl} & Halh Mongolian &	\langlabel{plt\_Latn} & Plateau Malagasy &	\langlabel{vec\_Latn} & Venetian \\
\langlabel{dzo\_Tibt} & Dzongkha &	\langlabel{khm\_Khmr} & Khmer (Central Khmer) &	\langlabel{pol\_Latn} & Polish &	\langlabel{vie\_Latn} & Vietnamese \\
\langlabel{ekk\_Latn} & Standard Estonian &	\langlabel{kik\_Latn} & Kikuyu (Gikuyu) &	\langlabel{por\_Latn} & Portuguese &	\langlabel{war\_Latn} & Waray (Philippines) \\
\langlabel{ell\_Grek} & Modern Greek (1453-) &	\langlabel{kin\_Latn} & Kinyarwanda &	\langlabel{quy\_Latn} & Ayacucho Quechua &	\langlabel{wol\_Latn} & Wolof \\
\langlabel{eng\_Latn} & English &	\langlabel{kir\_Cyrl} & Kirghiz (Kyrgyz) &	\langlabel{ron\_Latn} & Romanian (Moldavian, Moldovan) &	\langlabel{xho\_Latn} & Xhosa \\
\langlabel{epo\_Latn} & Esperanto &	\langlabel{kmb\_Latn} & Kimbundu &	\langlabel{run\_Latn} & Rundi &	\langlabel{ydd\_Hebr} & Eastern Yiddish \\
\langlabel{eus\_Latn} & Basque &	\langlabel{kmr\_Latn} & Northern Kurdish &	\langlabel{rus\_Cyrl} & Russian &	\langlabel{yor\_Latn} & Yoruba \\
\langlabel{ewe\_Latn} & Ewe &	\langlabel{knc\_Arab} & Central Kanuri &	\langlabel{sag\_Latn} & Sango &	\langlabel{yue\_Hant} & Yue Chinese \\
\langlabel{fao\_Latn} & Faroese &	\langlabel{knc\_Latn} & Central Kanuri &	\langlabel{san\_Deva} & Sanskrit &	\langlabel{zgh\_Tfng} & Standard Moroccan Tamazight \\
\langlabel{fas\_Arab} & Persian &	\langlabel{kor\_Hang} & Korean &	\langlabel{sat\_Olck} & Santali &	\langlabel{zsm\_Latn} & Standard Malay \\
\langlabel{fij\_Latn} & Fijian &	\langlabel{ktu\_Latn} & Kituba (Democratic Republic of Congo) &	\langlabel{scn\_Latn} & Sicilian &	\langlabel{zul\_Latn} & Zulu \\
\langlabel{fil\_Latn} & Filipino (Pilipino) &	\langlabel{lao\_Laoo} & Lao &	\langlabel{shn\_Mymr} & Shan &	\langlabel{zxx\_Zxxx} & No linguistic content (Not applicable) \\
\langlabel{fin\_Latn} & Finnish &	\langlabel{lat\_Latn} & Latin &	\langlabel{sin\_Sinh} & Sinhala (Sinhalese) &	& \\
    \bottomrule
\end{tabular}
}
\caption{OpenLID-v3 supported languages (model class labels).}
\label{tab:openlidv3-langs}
\end{table*}

\end{document}